\documentclass{article}

\usepackage{PRIMEarxiv}
\usepackage{amsmath}
\usepackage[utf8]{inputenc} 
\usepackage[T1]{fontenc}    
\usepackage{hyperref}       
\usepackage{url}            
\usepackage{booktabs}       
\usepackage{amsfonts}       
\usepackage{nicefrac}       
\usepackage{microtype}      
\usepackage{lipsum}
\usepackage{algpseudocode}
\usepackage{fancyhdr}       
\usepackage{graphicx}       
\graphicspath{{media/}}     
\usepackage{lscape}

\usepackage{subcaption}

\title{Edge2Node: Reducing Edge Prediction to Node Classification
}

\author{
  Zahed Rahmati\thanks{zrahmati@aut.ac.ir}\\Department of Mathematics and Computer Science\\Amirkabir University of Technology 
}

\begin{document}
\maketitle

\begin{abstract}
Despite the success of graph neural network models in node classification, edge prediction (the task of predicting missing or potential links between nodes in a graph) remains a challenging problem for these models.
A common approach for edge prediction is to first obtain the embeddings of two nodes, and then a predefined scoring function is used to predict the existence of an edge between the two nodes.  
Here, we introduce a preliminary idea called Edge2Node which suggests to directly obtain an embedding for each edge, without the need for a scoring function. This idea wants to create a new graph $H$ based on the graph $G$ given for the edge prediction task, and then suggests reducing the edge prediction task on $G$ to a node classification task on $H$. We anticipate that this introductory method could stimulate further investigations for edge prediction task.
\end{abstract}


\section{Introduction}
Given a graph $G= (V, E)$, where $V$ is the set of nodes and $E \subseteq V \times V$ is the set of edges, let $Y$ be the node label vector, where each component $y_v$ denotes the label of $v$, and let $X \in \mathbb{R}^{|V| \times d}$ be the node feature matrix, where each row $X_v$ corresponds to a $d$-dimensional feature vector for node $v$. 

Edge prediction is the task of predicting whether an edge exists between two nodes in a graph, given the features of the nodes and their neighbors.
Edge prediction has many applications, such as predicting future friendships or connections between users in social networks, suggesting potential links between users and products or content in recommendation systems, and anticipating undiscovered protein-protein interactions or potential disease associations in biological networks. Node classification is the task of predicting the label of a node in a graph, which is the most popular task in machine learning on graph-based data.

Graph Neural Networks (GNNs) are neural networks that can learn from and predict patterns in graph data. GNNs are able to capture the structural information of graphs, which makes them ideal for predicting potential connections between nodes. Graph neural networks (GNNs) have seen significant advancements in recent years, and they have been extensively studied for tasks like node and graph classification. However, their application in edge prediction has received comparatively less attention, offering substantial research opportunities; edge prediction is a more challenging task, as it requires the model to learn the complex relationships between nodes in a graph.

There are various methods and approaches for edge prediction tasks. Three primary categories of GNN-based edge prediction techniques include: Graph Autoencoder-Based (GAE) methods~\cite{kipf2016variational}, subgraph neural graph networks~\cite{zhang2018link}, and techniques that utilize neighborhood node embeddings after Message Passing Neural Network (MPNN) operations~\cite{gilmer2017neural}. Next we will discuss each.

Graph autoencoders (GAEs) are a powerful and flexible way to predict missing or future edges in a graph. GAEs work by training a neural network to learn a latent representation of the graph, which is then used to reconstruct the graph; \emph{i.e.,} it trains a graph autoencoder on the input graph to generate node embeddings for all nodes. The edge prediction task is then performed by predicting the edges in the reconstructed graph.
GAE cannot predict the dependency between two endpoints of an edge, so it fails in predicting edges in simple scenarios; see~\cite{zhang2021labeling} for more details.

To address issues raised by GAE methods, subgraph neural network methods have emerged as a promising new approach to edge prediction. These methods work by learning representations of subgraphs, which are small neighborhoods of nodes in the graph. The representations of the subgraphs are then used to predict the existence of edges between the nodes. One of the most successful subgraph neural network methods for edge prediction is SEAL (Learning from Subgraphs, Embeddings, and Attributes for Link Prediction)~\cite{zhang2018link}.

While learning-from-subgraph approaches have achieved impressive results on standard datasets, they also have some drawbacks. Calculating the subgraph for each edge is computationally expensive and time-consuming, especially for large networks and graphs. Additionally, the size of these subgraphs varies, which can reduce the efficiency of GPU-based training.
Researchers have introduced a number of innovative approaches to address the computational and time costs challenges of learning-from-subgraph approaches for edge prediction; see~\cite{chamberlain2023graph},~\cite{yun2021neo}, and~\cite{wang2023neural}.

Traditional methods for edge prediction first learn node representations using a message passing algorithm. Then, they use a predefined scoring function (\emph{e.g.,} the dot product of the two node embeddings, or a more complex one such as a logistic regression model)  to predict the existence of an edge between two nodes based on their representations.

Here, we propose a preliminary approach, called Edge2Node, which tries learning a direct embedding for the edge whose existence wanted to be checked. Edge2Node takes the input graph for the edge prediction task and constructs a new graph for a node classification task, whose predicted labels are equivalent to the existence of edges in the input graph. Edge2Node in fact is trying to fully utilize the power of MPNNs by exploiting all of their structural features and advantages.

\section{Edge Prediction to Node Classification}
\label{sec:headings}
In this section, we construct a new graph $H$ based on the input graph $G=(E, V)$, and then we show how the edge prediction task on $G$ can be reduced to a node classification task on $H$. 

Negative edge sampling is essential for edge prediction because it enables the model to learn the distinction between positive and negative edges. Therefore, we use negative edge sampling techniques from the litrature~\cite{perozzi2014deepwalk,grover2016node2vec,yang2020understanding} to add edges between nodes that are not connected in the original graph $G$. 

Denote the set of all new, negative edges by $E'$, and denote the graph constructed by adding all the edges of $E'$ to the graph $G$ by $G'=(E\cup E', V)$. For the negative edges, we randomly sample $|E'|$ pairs of nodes from the original graph $G$. $|E'|$ is a hyperparameter depending on the input graph and the training edge size, which is usually one third of $|E|$, the size of the original (positive) edges in $G$.

\subsection{Adding Dummy Nodes and Edges}
\begin{figure}[h!]
	\centering
	\includegraphics[scale=1.2]{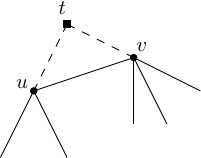}
	\caption{Adding a dummy node $t$ and its incident dummy edges $(t,u)$ and $(t,v)$.}
 \label{fig:dummy}
	\end{figure}

For any edge $(u,v)$ of the graph $G'$ , we create a  dummy node $t$ and add two incident dummy edges $(t, u)$ and $(t, v)$; see Figure~\ref{fig:dummy}. Let $V'$ be the set of all dummy nodes, and let $E''$ be the set of all  dummy edges. 
We denote by $H=(E\cup E'\cup E'', V\cup V')$ the graph which is constructed by the edge set $E\cup E'\cup E''$ on the node set $V\cup V'$.

\subsection{Reducing Edge Prediction to Node Classification}\label{algo}
Here, we first label the nodes of $H$. We set the label $y_w$ of the original nodes $w\in V$ of $H$ to 0, and set $y_w$ to 1 if $w$ corresponds to a positive (original) edge and set $y_w$ to 2 if $w$ corresponds to a negative edge, \emph{i.e.,}

\[y_w= \left\{ \begin{array}{cl}
0 & : \ \text{if $w\in V$} \\
1 & : \ \text{if $w\in V'$ corresponds to an edge in $E$} \\
2 & : \ \text{if $w\in V'$ corresponds to an edge in $E'$}.
\end{array} \right.\]

Using this mapping, the original edge prediction task on $G$ is now equivalent to a node classification task on $H$. The probability of existence of a positive edge $e$ is equal to the probability of $y_w=1$, where $w$ corresponds to the edge $e$ we needed to predict for.

\begin{figure}[h!]
  \centering
    \begin{subfigure}[b]{0.3\linewidth}
    \includegraphics[width=\linewidth]{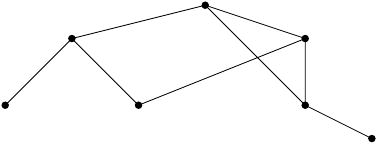}
    \caption{Original Graph $G$}
  \end{subfigure}
  \hspace{5mm}
    \begin{subfigure}[b]{0.35\linewidth}
    \includegraphics[width=\linewidth]{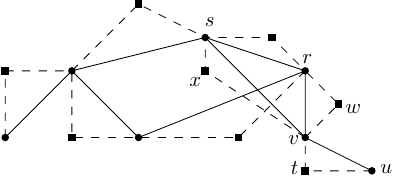}
    \caption{The new Graph $H$}
  \end{subfigure}
  \hspace{5mm}
    \begin{subfigure}[b]{0.25\linewidth}
    \includegraphics[width=\linewidth]{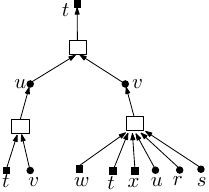}
    \caption{GNN Model}
  \end{subfigure}
  \caption{After augmenting the original graph $G$, by adding dummy nodes and dummy edges, a GNN model on $H$ can be trained to do the node classification.}
  \label{fig:coffee}
\end{figure}

We can now feed the new graph $H$ to our desired GNN model, \emph{e.g.,} GCN and GraphSAGE~\cite{kipf2016semi,hamilton2017inductive}, and use the final embeddings of the dummy nodes for classification, which is equivalent to predicting the existence of the corresponding edges. Figure~\ref{fig:coffee} depicts an overview of our Edge2Node approach.


\section{Conclusion}
This paper tackled the challenging problem of predicting edges in a graph, a task that has traditionally been difficult for both computational algorithms and machine learning methods.
MPNNs, which work by passing messages between neighboring nodes in a graph to learn representations that are invariant to graph permutations, are typically used for node classification. Our preliminary Edge2Node approach, which suggests doing the edge prediction task via a node classification task, wants to fully utilize the power of MPNNs by exploiting this ability to learn robust and informative representations for edges.

We believe that this preliminary approach would help to obtain new directions for edge prediction tasks.


\bibliographystyle{unsrt}  
\bibliography{references}

\end{document}